\documentclass[acmtog]{acmart}
\newcommand{\AD}{Animated Drawings }
\newcommand{\acknowledgement[2]}{{#2}{}}

\newcommand{\hjs[2]}{{#2}{}}

\usepackage{censor}

\AtBeginDocument{%
  \providecommand\BibTeX{{%
    \normalfont B\kern-0.5em{\scshape i\kern-0.25em b}\kern-0.8em\TeX}}}

\setcopyright{acmcopyright}
\copyrightyear{2023}
\acmYear{2023}
\acmBooktitle{}

\acmSubmissionID{0115}

\usepackage{soul}

\begin{document}

\title{A Method for Animating Children's Drawings of the Human Figure}


\author{Harrison Jesse Smith}
\affiliation{%
  \institution{Meta AI Research}
  \country{USA}
}
\email{hjessmith@gmail.com}

\author{Qingyuan Zheng}
\affiliation{%
  \institution{Tencent America}
  \country{USA}
}
\email{qyzzheng@global.tencent.com}
\authornote{Author was affiliated with Meta AI Research while contributing to this work.}

\author{Yifei Li}
\affiliation{%
  \institution{MIT CSAIL}
  \country{USA}
}
\email{liyifei@csail.mit.edu}
\authornotemark[1]

\author{Somya Jain}
\affiliation{%
  \institution{Meta AI Research}
  \country{USA}
}
\email{somyaj@gmail.com}

\author{Jessica K. Hodgins}
\affiliation{%
  \institution{Carnegie Mellon University}
  \country{USA}
}
\email{jkh@cmu.edu}
\authornotemark[1]

\renewcommand{\shortauthors}{Smith et al.}

\begin{abstract}
Children's drawings have a wonderful inventiveness, creativity, and variety to them.
We present a system that automatically animates children's drawings of the human figure, is robust to the variance inherent in these depictions, and is simple and straightforward enough for anyone to use. 
We demonstrate the value and broad appeal of our approach by building and releasing the \AD Demo, a freely available public website that has been used by millions of people around the world.
We present a set of experiments exploring the amount of training data needed for fine-tuning, as well as a perceptual study demonstrating the appeal of a novel \textit{twisted perspective} retargeting technique. 
Finally, we introduce the Amateur Drawings Dataset, a first-of-its-kind \hjs{annotated} dataset, collected via the public demo, containing over 178,000 amateur drawings and corresponding \hjs{user-accepted} character bounding boxes, segmentation masks, and joint location annotations.
\end{abstract}




\begin{teaserfigure}
  \includegraphics[width=\textwidth]{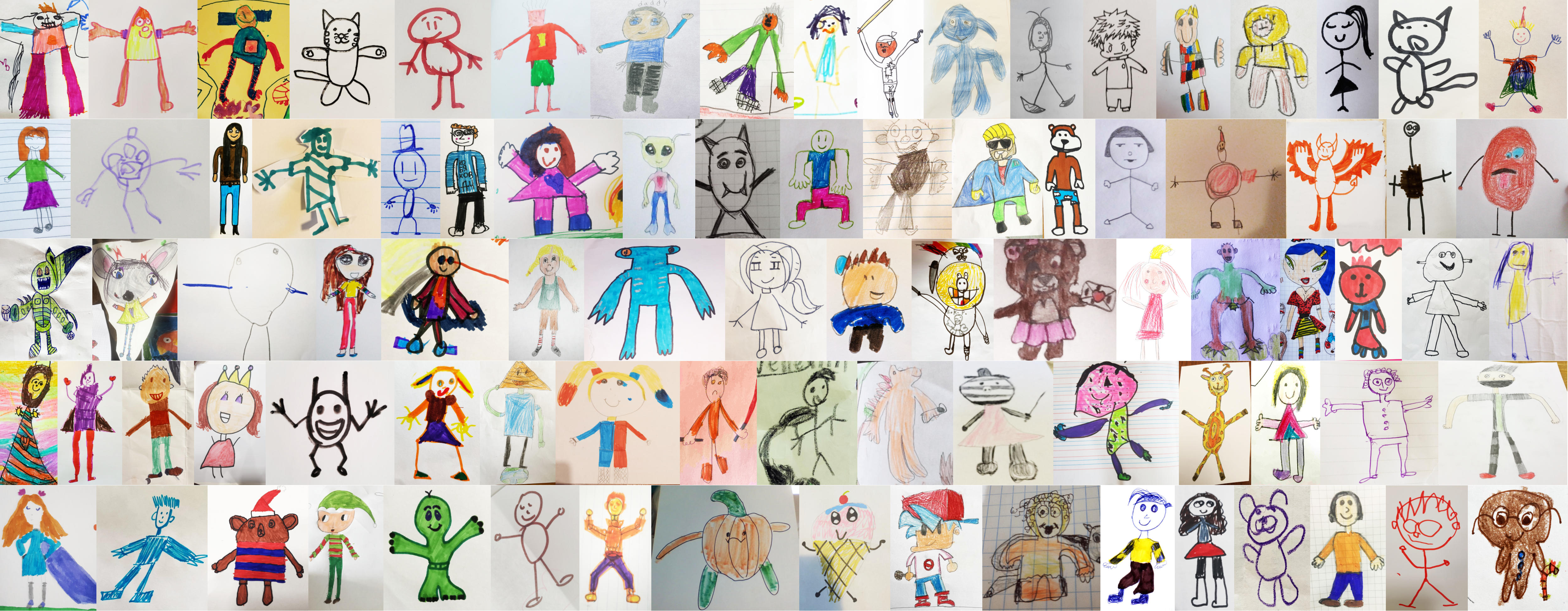}
  \caption{\hjs{We present a fast and easy-to-use method for animating the types of abstract and varied human-like figures drawn by children.}}
  \Description{Banner Caption}
  \label{fig:teaser}
\end{teaserfigure}

\maketitle

\section{Introduction}
Children's depictions of the human figure are highly expressive and varied.
As one of the very first subjects children attempt to draw, these representations begin as an almost unintelligible cloud of scribbles. 
As a child grows, their representation of the human figure becomes more developed and is extended to graphically represent many different types of characters: people, animals, and even personified objects (see Figure 1).

Who among us has not wished, either as a child or as an adult, to see such figures come to life and move around on the page?
Sadly, while it is relatively fast to produce a single drawing, creating the sequence of images necessary for animation is a much more tedious endeavor, requiring discipline, skill, patience, and sometimes complicated software.
As a result, most of these figures remain static upon the page.

Inspired by the importance and appeal of the drawn human figure, we design and build a system to automatically animate it given an in-the-wild photograph of a child's drawing. 
Our system is fast, intuitive, and robust to much of the variation present in these types of drawings, making it well-suited to allow our target audience--children--to see their own characters coming to life.
The system is comprised of four stages: figure detection, segmentation masking, pose estimation/rigging, and animation. 
We describe each stage and identify common causes of failure in each. 
For object detection and pose estimation, we make use of existing computer vision models designed to detect human figures and joints in photographs; we fine-tune these models for use with children's drawings.
For segmentation, we present a straightforward, image processing-based method that, for animation purposes, is more useful and accurate than segmentation masks obtained from a fine-tuned object detection model.
During the animation step, we take advantage of the \textit{twisted perspective} commonly seen in children’s drawings to retarget motion capture data onto the character in a novel and appealing way.

While our system leverages existing models and techniques, most are not directly applicable to the task due to the many differences between photographic images and simple pen and paper representations. 
Therefore, we couple the presentation of our system with a set of experiments exploring the relationship between fine-tuning training set size and success rates.
We also include a perceptual study validating viewer preference for incorporating \textit{twisted perspective} into the motion retargeting step.

We validate the desirability and appeal of our system by building and publicly releasing a version of it as the \AD Demo\,\cite{animateddrawings}.
Launched in December 2021, this demo has been used by millions of people around the world to animate their drawings.
Inspired by this reception, our second contribution is the Amateur Drawings Dataset: \hjs{178,166 drawings and user-accepted annotations collected, with consent, through the demo. See Section \ref{sec:UI} for a description of how the annotations were generated.}
We believe this dataset will be a resource to researchers from various fields seeking to better understand the space of amateur drawings, evaluate new algorithms in this domain, or develop new drawing-based tools in general.

To summarize, our contributions are as follows:
\begin{enumerate}
    \item 
    We explore the problem of automatic sketch-to-animation for children's drawings of human figures and present a framework that achieves this effect. We also present a set of experiments determining the amount of training data necessary to achieve high levels of success and a perceptual study validating the usefulness of our motion retargeting technique.
    \item To encourage additional research in the domain of amateur drawings, we present a first-of-its-kind dataset of 178,166 user-submitted amateur drawings, along with user-accepted bounding box, segmentation mask, and joint location annotations.
\end{enumerate}

In addition, we also provide the project's animation code and the fine-tuned model weights for drawn human figure detection and pose estimation.

\section{Background}
Our work builds on existing methods from several fields but is, to our knowledge, the first work focused specifically on fully automatic animation of children's drawings of human figures. 
To ground the work, we provide a summary of salient observations from the field of children's art analysis.
In addition, we briefly review related work on 2D image-to-animation and object and pose estimation for abstract images.

\subsection{Analysis of Children's Drawings}

\hjs{
Children's drawings have long been of interest to the scientific community.
For well over a century, researchers from multiple fields have 
collected\,\cite{IndianaS55:online,kellogg1967rhoda,AWebbasedDatabaseforDrawingsofGods,geist2002they}
and analyzed them, seeking to understand and measure children's thought processes\,\cite{sully2021studies,barnes1892study,clark1897child,buhler2013mental}, 
intellectual development\,\cite{goodenough1926measurement},
and perceptions\,\cite{chambers1983stereotypic,doi:10.1080/01443410500344167}.
}
Particular attention has been given to drawings of human figures, one of the first and most frequently drawn subjects throughout childhood\,\cite{cox2013children}.

\begin{figure}
\includegraphics[width=\linewidth]{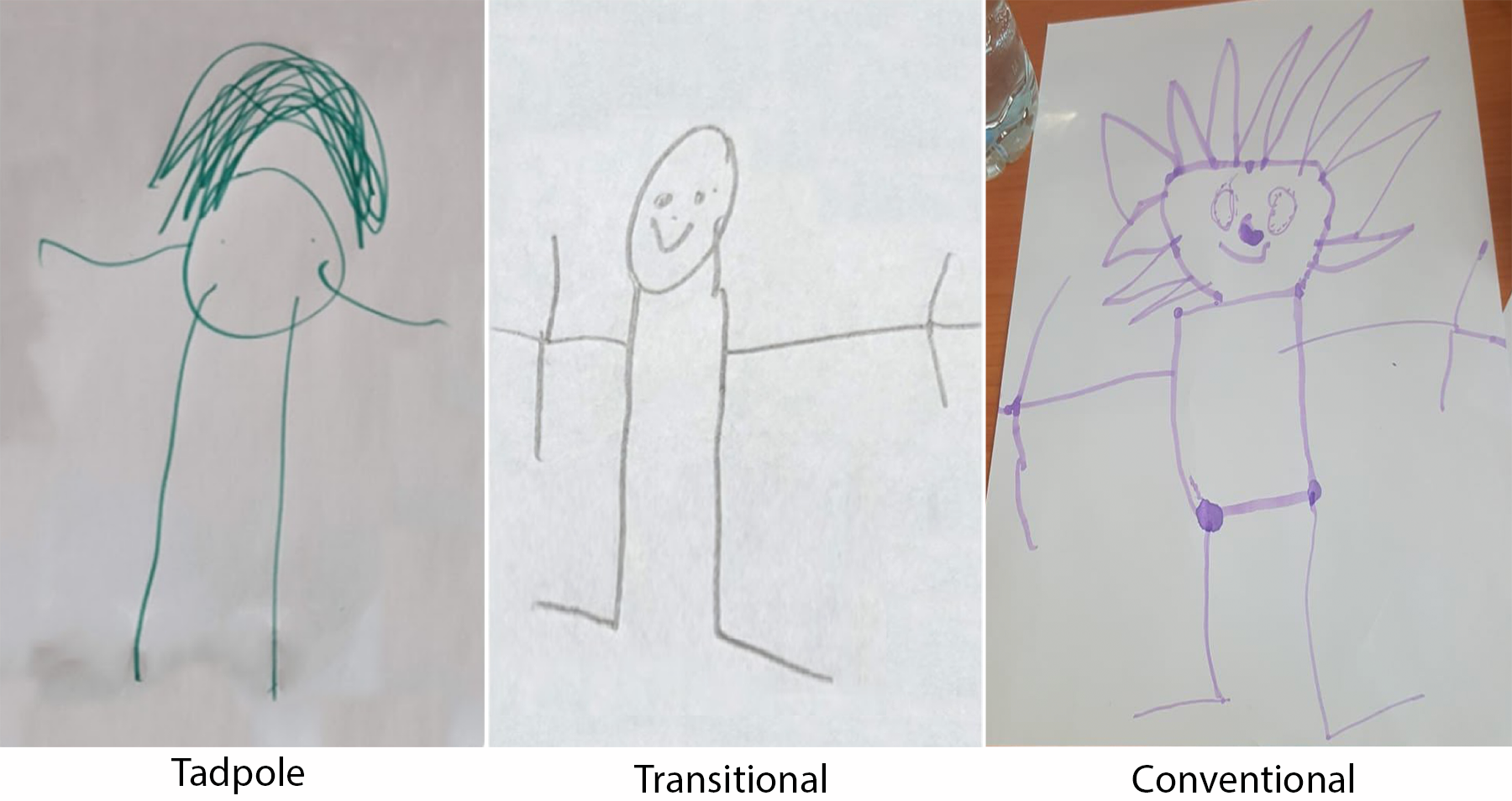}
\caption{
As children learn to draw the human figure, the morphologies of the schemas they employ vary and evolve considerably\,\cite{cox2014drawings}.
Children frequently begin by drawing a \textit{tadpole figure}, a circular head region from which arms and legs extend. 
Some will progress to a \textit{transitional figure}, dropping the arms down so they extend from the legs. 
When a line is drawn between the legs, creating the separate torso region, the \textit{conventional figure} is formed.
Though these are small changes from the perspective of the drawer, they result in significantly different character morphologies when viewed through the lens of character animation.
A successful drawing-to-animation system must be robust to these types of variations.}
\label{fig:tadpole-transitional-conventional}
\end{figure}

As the child develops, the schemas they employ to represent the human form become more complete (see Figure \ref{fig:tadpole-transitional-conventional}).
Even within these schemas, there is significant variation.
In addition to asymmetries and variation in color and proportion, many body parts appear optional to include; a study of drawings by 4-6 year old children showed that, while heads, legs, and eyes are almost universally present, other body parts (including torsos, arms, hands, and feet) were frequently absent\,\cite{cox2013children}.
Inversely, non-human body parts are frequently added in order to represent other subject classes\,\cite{kellogg1969analyzing}. With the addition of large ears, the figure may represent a cat or bear (Figures \ref{fig:maskrcnn_before_after}.m and \ref{fig:maskrcnn_before_after}.g); with the addition of a crown, it can represent a pineapple (Figure \ref{fig:maskrcnn_before_after}.n).
All of these sources of character variation make automatic character animation from drawings a non-trivial task.

Many researchers have focused closely on the unique style of children's drawings.
The psychologist and artist John Willats argues that, in order to understand the style of children's drawings, one must understand that the primary picture primitives employed by children are \textit{regions}, or 2D areas\,\cite{willats2006making}.
A squat volume, such as a head or torso, may be represented by a circular or ellipsoid region, whereas an elongated volume, such as a leg, may be represented by a long, thin region or even a single line.
These regions are not depictions of the object from any particular point of view. 
Rather, they are \textit{3D volumetric object-centered descriptions}\,\cite{marr1982vision},
2D areas with attributes perceptually similar to those of 3D object they are meant to represent.

There are two stylistic outcomes of these \textit{object-centered descriptions} that bear mention.
First, the use of foreshortening is very rare in children's drawings \,\cite{piaget1956, willats1992representation}. 
This design choice is understandable; foreshortening a long region, such as a limb, results in a short region which does not adequately reflect the \textit{longness} of the object.
Second, the human figure may appear to have been drawn from many different perspectives, so as to make each part of the character maximally recognizable.
For example, the head and torso may face forward while the legs and feet are pointed to the side.
This technique, often referred to as \textit{twisted perspective}, is frequently seen and well-documented\,\cite{dziurawiec1992twisted}.
Both of these stylistic aspects are used to guide the design decisions of our system when applying human motion capture data onto the character.

\subsection{2D Image to Animation}

Previous researchers have proposed methods to animate drawings or photographs, many of which rely upon additional modes of user input.
Hornung et al. present a method to animate a 2D character in a photograph, given user-annotated joint locations\,\cite{Hornung2007anim2Dpicmotion}.
Pan and Zhang demonstrate a method to animate 2D characters with user-annotated joint locations via a variable-length needle model\,\cite{Pan2011}.
Jain et al. present an integrated approach to generate 3D proxies for animation given joint locations, segmentation masks, and per-part bounding boxes\,\cite{jain:2012}. 
Levi and Gotsman provide a method to create an articulated 3D object from a set of annotated 2D images and an initial 3D skeletal pose\,\cite{ArtiSketch}.
\textit{Live Sketch}\,\cite{su2018livesketch}
tracks control points from a video and applies their motion to user-specified control points upon a character.
Other approaches allow the user to specify character motions through a puppeteer interface, using RGB or RGB-D cameras\,\cite{held20123d,barnes2008video}.
\textit{ToonCap}\,\cite{Fan:2018:TAL} focuses on an inverse problem, capturing poses of a known cartoon character, given a previous image of the character annotated with layers, joints, and handles.

\textit{Toonsynth}\,\cite{Dvoroznak18-SIG} and \textit{Neural Puppet}\,\cite{poursaeed2020neural} both present methods to synthesize animations of hand-drawn characters given a small set of drawings of the character in specified poses.
Hinz et al. train a network to generate new animation frames of a single character given 8-15 training images with user-specified keypoint annotations\,\cite{hinz2022charactergan}.

\textit{Monster Mash}\,\cite{Dvoroznak20-SA} presents an intuitive framework for sketch-based modeling and animation, and \textit{2.5D Cartoon Models}\,\cite{10.1145/1778765.1778796} presents a novel method of constructing 3D-like characters from a small number of 2D representations. 
Both of these are intuitive and well designed animation tools targeted towards amateur users.

\hjs{
Some animation methods are specifically tailored toward particular forms, such as faces\,\cite{elor2017bringingPortraits}, coloring book characters\,\cite{magnenat2015live}, or characters with human-like proportions. 
One notable work that is focused on the human form is \textit{Photo Wake Up}\,\cite{weng2019photo}. 
The authors show a method for creating a rigged and textured 3D mesh from a single image of a human-like figure.
Similar to us, their end goal is to allow users to seamlessly bring 2D characters to life; their work does an impressive job of accomplishing this.
Our method differs in two significant ways. 
First, while their work is focused on creating a 3D model for a mixed reality use case, 
ours is specifically focused on animating twisted perspective figures while staying within a 2D plane.
Second, children's drawings are much more abstract, incorrectly proportioned, and non human-like than the examples demonstrated in the paper.
We test our method upon the more abstract examples demonstrated in their paper and, with minor segmentation cleaning, they were successfully animated by our method.
}

\hjs{While the approaches listed here are wonderful tools to ease the burden of animation, none were perfectly suited to our use case.
Some require additional user input beyond the drawing itself, making the animation process more complex.
Others require the user to consistently draw the same character in multiple poses, which is beyond the skills of young children.
Others are focused on animating specific forms, precluding their use on children's drawings of the human figure.}


\subsection{Detection, Segmentation, and Pose Estimation on Non-Photorealistic Images}

\hjs{
Aided by the the existence of large annotated datasets\,\cite{lin2014microsoft,6909866,6682899}, researchers have made considerable progress solving the problems of object detection, segmentation, and pose estimation from photographs. See, for example\,\cite{MaskRCNNhe2017mask,openpose19,guler2018densepose,alphapose,toshev2014deeppose}.
We explain the methods in this area that we leverage in Sections \ref{sec:character_detection} and \ref{sec:joint_detection}.

While traditional methods for detection, segmentation, and pose estimation of non-photorealistic images exist\,\cite{choi2012retrieval,bregler2002turning,davis2006sketching,eitz2012humans}, the lack of easily available datasets has resulted in slower adoption of deep learning models.
Some researchers are addressing this problem by developing methods and releasing datasets focused on the domain of anime characters\,\cite{chen2022bizarre,10.1145/3011549.3011552}, professional sketches\,\cite{brodt2022sketch2pose}, and mouse doodles\,\cite{ha2017neural}.
Other researchers have presented a non-deep learning method for inferring character poses from \textit{gesture drawings}\,\cite{Gesture3D}.
}
Because the Amateur Drawings Dataset is comprised of in-the-wild photographs of drawings created by the general public, we believe it will complement the value of existing datasets and allow for new dimensions of exploration and analysis.

\section{Method}

\begin{figure*}[ht]
  \centering
  \includegraphics[width=\linewidth]{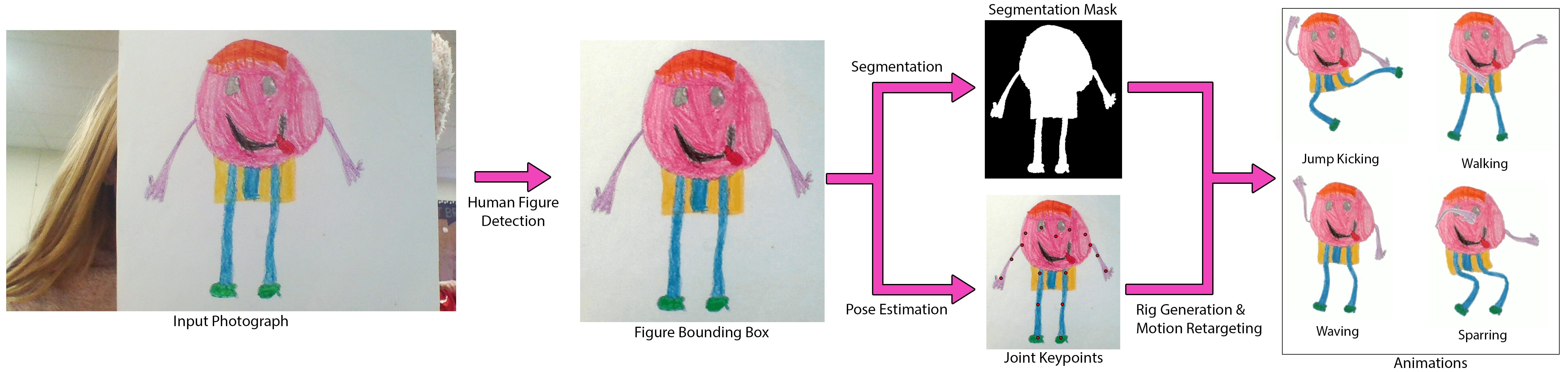}
  \caption{An overview of the drawing-to-animation pipeline. Given an input drawing, the human figure within it is identified and used to crop the image. From the cropped image, the human figure segmentation mask and joint locations are obtained and used to create a character rig. Motion capture data is then retargeted onto the character rig to produce animations.}
  \Description{Overview Stand In.}
  \label{fig:overview}
\end{figure*}

Our goal is a system that generates an animation from a single drawing of a human figure.
To make the experience as simple and accessible as possible, we take the input to be a single in-the-wild photograph of a drawing, as might be captured with a mobile phone camera.
While vector drawings from tablet-based interfaces can provide stroke-level information, previous in-classroom research has found tablet-based drawing interfaces to be more fatiguing and difficult to use than their analog counterparts\,\cite{picard2014ipads}; we therefore assume the input to be a raster image that is a photograph of a child's drawing with pen, crayon, paint, or other common art material.

Starting with the single image, we structure the task as a series of sub-tasks: human figure  detection, segmentation, pose estimation, and animation (Figure \ref{fig:overview}).
The first step is to identify the human figure within the drawing and predict a bounding box that tightly encompasses it.
Second, we use the contents of the bounding box to obtain a segmentation mask, separating pixels belonging to the human figure from those belonging to the background.
Third, we use the contents of the bounding box to perform pose estimation on the figure, identifying a series of skeletal joints.
With the original image, segmentation mask, and joint locations, we generate a character rig suitable for animation.
Finally, we animate the character rig by retargeting motion capture data onto it.

In the following sections, we describe the steps in more detail and provide examples of common failures that can occur.
We end by describing how the system is framed within the publicly released \AD Demo and how the user interface is structured to allow users to modify the model predictions as needed.

\subsection{Figure Detection}
\label{sec:character_detection}
We first detect a bounding box around the human figure within the drawing.
This step is necessary because many children's drawings portray human figures as part of a larger scene \cite{kellogg1967rhoda} and because the photograph may include background  either drawn or outside the bounds of the piece of paper such as a table surface.

We make use of a state-of-the-art object detection model, Mask R-CNN\,\cite{MaskRCNNhe2017mask}, with a ResNet-50+FPN backbone.
We utilize pretrained weights derived from the MS-COCO dataset, one of the largest publicly available semantic segmentation datasets\,\cite{lin2014microsoft}. 
However, MS-COCO is comprised primarily of photographs of real-world objects, not artistic renderings, and does not contain a category for \textit{drawings of human figures}. Therefore, we fine-tune the model.
The model's backbone weights are frozen and attached to a head, which predicts a single class, \textit{human figure}. 
The weights of the head are then optimized using cross-entropy loss and stochastic gradient descent with an initial learning rate of 0.02, momentum of 0.9, weight decay of 1e-4, and minibatches of size 8.
Training was conducted using OpenMMLab Detection Toolbox \cite{mmdetection}; all other hyperparameters were kept at the default values provided by the toolbox.
Each model was trained until convergence on a server with eight Tesla V100-SXM2 GPUs.

In Figure \ref{fig:maskrcnn_before_after}, we show representative example predictions. 
See supplemental material for additional examples.
For an exploration of the amount of training data necessary to achieve acceptable results, see Section \ref{sec:effect_of_training_sample_size}.

\begin{figure*}[ht]
  \centering
  \includegraphics[width=\linewidth]{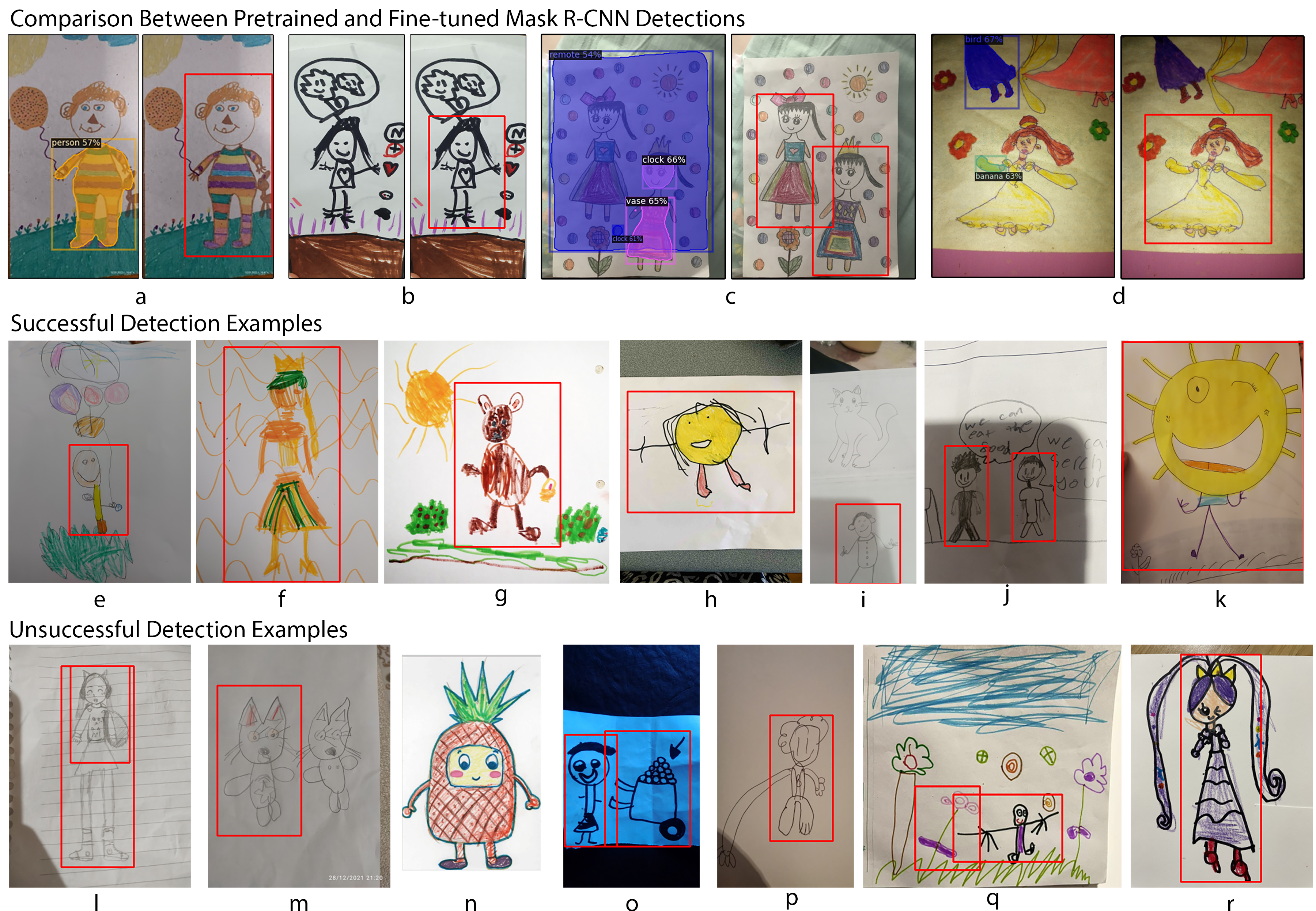}
  \caption{
  Row 1 shows representative detection failures from pretrained Mask R-CNN (left) that were corrected after model fine-tuning (right): 
    excluding hollow parts of a figure (a),
    false negatives (b),
    incorrectly detecting objects in the background (c),
    and detecting and incorrectly classifying parts of figures (c, d).
  Row 2 contains examples of successful detections from the fine-tuned model. 
  Row 3 contains representative examples of failures: 
    multiple detections of the same figure (l),
    false negatives (m, n),
    false positives (o, q),
    and detections that cut off figure parts (p, r).
    Additional examples are shown in the supplemental material.
}
\Description{Mask R-CNN before and after fine-tuning.}
\label{fig:maskrcnn_before_after}
  
\end{figure*}

\subsection{Figure Segmentation}
\label{sec:character_segmentation}
With the bounding box identified, we next obtain a segmentation mask, separating the figure from the background.
This step is surprisingly difficult; there is a great deal of variation in figure appearance and in photograph quality.
Additionally, texture and color, two attributes that are useful for segmentation in photographs, are of limited value here: they are a function of the artist's drawing style and their available drawing tools.
While Mask R-CNN does predict a segmentation mask for each detection, we found them to be inadequate in many cases. 
Because this mask will be used to create a 2D textured mesh of the figure, it must be a single polygon that tightly conforms to the edges of the figure, includes all body parts, and excludes extraneous background elements.

\begin{figure*}[ht]
  \centering
  \includegraphics[width=\linewidth]{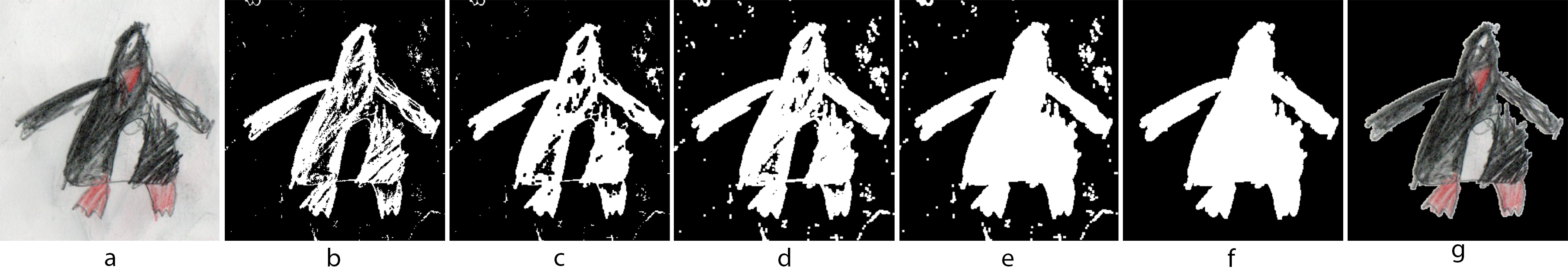}
  \caption{
We use an image processing-based approach to extract the figure mask. Beginning with the contents of the detected bounding box (a), we convert to grayscale and apply adaptive thresholding (b), perform morphological closing (c) and dilating (d), flood filling (e), and retain only the largest polygon (f). Here the resulting mask tightly conforms to the original figure (g). 
}
\Description{Segmentation Flowchart}
\label{fig:segmentation-flowchart}
  
\end{figure*}

We therefore use a classical, image processing-based approach for extracting masks (see Figure \ref{fig:segmentation-flowchart}). 
First, we resize the bounding box-cropped image to a width of 400 pixels while preserving the aspect ratio.
Next, we convert the image to grayscale and perform adaptive thresholding, where the threshold value is a Gaussian-weighted sum of the neighborhood pixel values minus a constant \textit{C} \cite{gonzalez2008digital}.
Here, we use a distance of 8 pixels to define the neighborhood and a value of 115 for \textit{C}.
To remove noise and connect foreground pixels, we next perform morphological closing, followed by dilating, using 3x3 rectangular kernels.
We then flood fill from the edges of the image, ensuring that any closed groups of foreground pixels are solid and do not contain holes.
Finally, we calculate the area of each distinct foreground polygon and retain only the one with the largest area.

\begin{figure*}[ht]
  \centering
  \includegraphics[width=\linewidth]{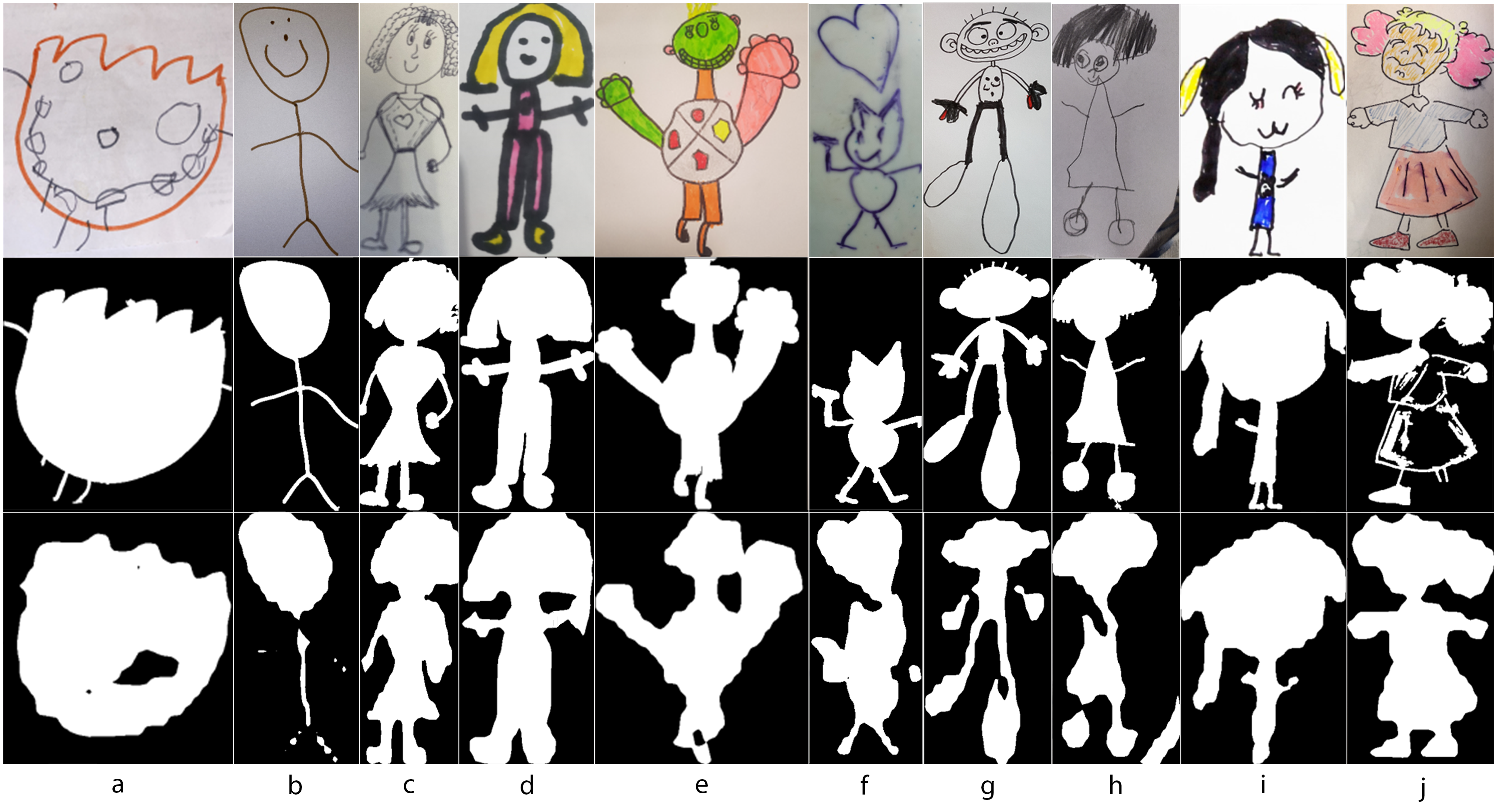}
  \caption{Given the input images cropped to the computed bounding boxes, shown in the top row, the image processing-based segmentation method computes the masks shown in the middle row. The bottom row shows the masks predicted by the fine-tuned Mask R-CNN model. 
  Often the image processing method gives usable results while the Mask R-CNN model excludes or detaches body parts (a, b, g, h), improperly attaches limbs to the body or head (c, d, e, f,) or includes non-figure elements (f, h).
  Columns i and j show examples in which the image processing method fails to extract a good mask, which can occur when the limbs of the figure are not drawn attached to the body (i) or the strokes outlining the figure are not connected (j). Note that (j) is an example of a figure for which the Mask R-CNN segmentation prediction is more suitable for animation than the mask obtained through our image processing-based segmentation method.
  }
  \Description{Segmentation Comparison}
  \label{fig:segmentation_comparison}
\end{figure*}

While this method is straightforward, we nonetheless found it to be an effective method for extracting useful and precise figure masks.
However, it will fail when body parts are drawn separated, limbs are drawn touching at points other than the joints, the figure is not fully contained by the bounding box, or the outline of the figure is not completely connected.
For examples comparing the Mask R-CNN segmentation predictions to the image-based processing approach, see Figure~\ref{fig:segmentation_comparison}.

\subsection{Pose Estimation}
\label{sec:joint_detection}

\begin{figure*}[ht]
  \centering
  \includegraphics[width=\linewidth]{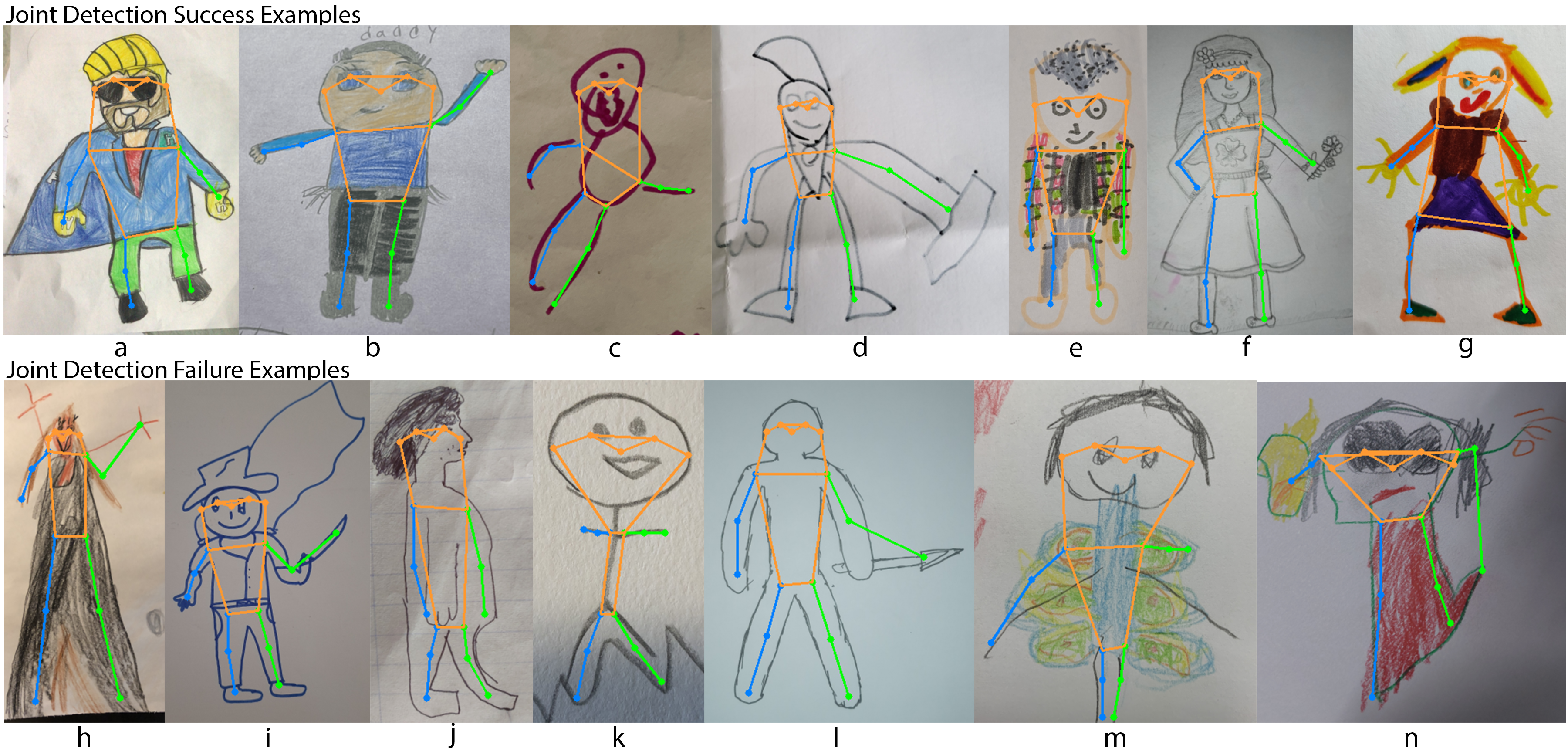}
  \caption{Examples of successful and unsuccessful pose estimations. Frequent causes of failure include limb confusion caused by background elements (k), limb confusion caused by other figure parts (h, m, n), and objects held by the figure (i, l). Human figures not drawn facing forward, while infrequent, also result in failure (j).
  Additional examples are shown in the supplemental material.
  }
  \Description{Examples of successful and unsuccessful pose estimations.}
  \label{fig:pose_examples}
  
\end{figure*}

\begin{figure*}[ht]
  \centering
  \includegraphics[width=\linewidth]{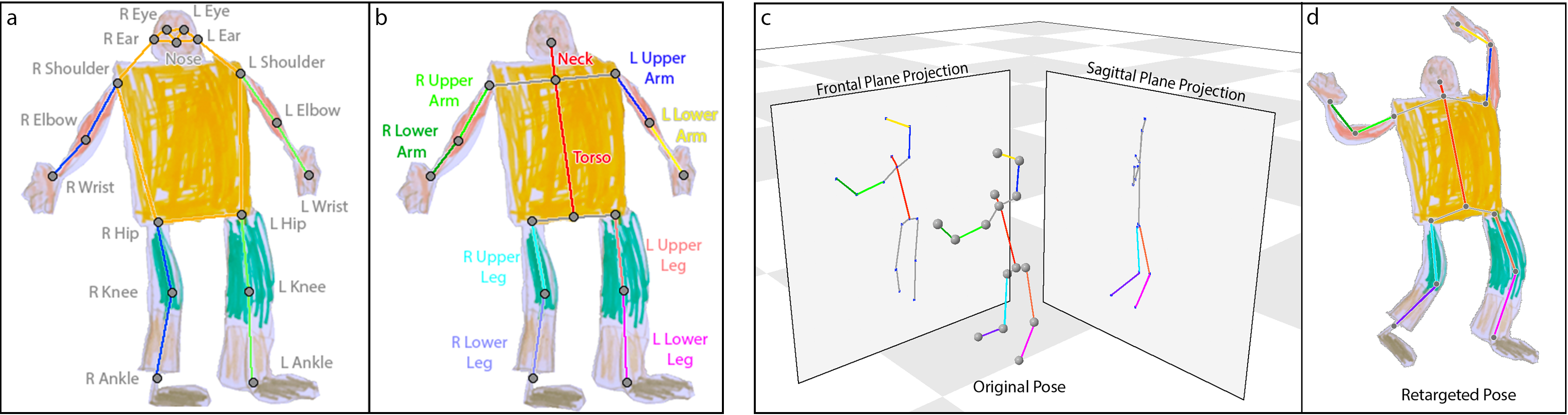}
  \caption{\textit{Left:} 
  Given the predicted joint keypoints (a), we create a skeletal rig used to animate the character (b).
  \textit{Right:} 
  In this example, we take the original pose from a motion capture actor and project the torso and upper limb joint locations onto a frontal plane, while projecting the lower limb joint locations onto a sagittal plane (c).
  We then find the global orientations of the bones within their respective planes and rotate the character's joints to match, resulting in the retargeted pose (d).
  }
  \Description{Rigged Character Creation}
  \label{fig:rigged_character_creation}
\end{figure*}

To allow the character to perform complex motions, we need an understanding of its proportions and pose.
However, a fine-grained analysis of a figure's body parts is tricky, due to the sparse and abstract way in which they can be represented;
a single line may be the edge of an arm (Figure \ref{fig:pose_examples}.l), an entire arm (Figure \ref{fig:pose_examples}.k), a design on the figure's shirt (Figure \ref{fig:pose_examples}.e), a background element (Figure \ref{fig:maskrcnn_before_after}.f), or a preexisting print upon the page (Figure \ref{fig:maskrcnn_before_after}.l),

Discerning exactly what each stroke of a drawing is can be difficult, even for humans.
To make this task more tractable, we instead only seek to identify a small set of keypoints that can be used as joints during the animation step.
We assume the presence of the 17 keypoints used by MS-COCO \cite{lin2014microsoft} (see Figure \ref{fig:rigged_character_creation}) and use a pose estimation model to predict their locations.

While there are many pose estimation models suitable for photographs of people, they do not perform well upon images of drawn human figures, which are quite different in appearance.
We therefore train a custom pose estimation model utilizing a ResNet-50 backbone, pretrained on ImageNet, and a top-down heat map keypoint head that predicts an individual heatmap for each joint location. 
The cropped human figure bounding box is resized to 192x256 and fed into the model, and the highest-valued pixel in each heatmap is taken as the predicted joint location.
Mean squared error is used for joint loss, and optimization is performed using Adaptive Momentum Estimation with learning rate of 5e-4 and minibatches of size 512.
Training was conducted using the OpenMMLab Pose Toolbox \cite{mmpose2020}; all other hyperparameters were kept at the default values provided by this toolbox.
The model was trained on a server with eight Tesla V100-SXM2 GPUs until convergence.

We provide representative examples of successful and unsuccessful pose estimation examples in Figure \ref{fig:pose_examples}.
See the supplemental material for additional examples.
As with detection, see Section \ref{sec:effect_of_training_sample_size} for an exploration of the amount of training data necessary to achieve acceptable results.

\subsection{Animation}

We next create a rigged character, suitable for animation, from the mask and joint predictions.
From the segmentation mask, we use Delaunay triangulation to generate a 2D mesh and texture it with the original drawing.
Using the joint locations, we construct a character skeleton.
We use the predicted positions of the left and right shoulders, elbows, wrists, hips, knees, ankles, and nose.
We average the position of the two hips to obtain a root joint and average the position of the two shoulders to obtain the chest joint.
We connect these joints to create the skeletal rig as shown in Figure \ref{fig:rigged_character_creation}.b.
Finally, we assign each mesh triangle to one of nine different body part groups (left upper leg, left lower leg, right upper leg, right lower leg, left upper arm, left lower arm, right upper arm, right lower arm, and trunk) by finding the closest bone to each triangle's centroid.
During the animation step, different body part groups can be rendered in different orders, giving the illusion of limbs being in front of or behind the body.

We animate the character rig by translating the joints and using as-rigid-as-possible (ARAP) shape manipulation\,\cite{igarashi2005asrigidaspossible} to repose the character mesh. 
To make the process simple for the user, we drive the character rig using a library of preselected motion clips obtained from human performers.
Because the human figures are 2D and often have very different proportions and appearances from those of real humans, care must be taken when deciding how to best utilize the 3D motion data.
We retarget the motion in the following manner.

We initially preprocess a motion clip by subtracting, per frame, the X and Z position of the root joint from the motion caption actor's skeleton, such that the skeleton's root joint is always located above the origin. 
We also rotate the skeleton about the vertical axis such that its forward vector (defined as the vector perpendicular to the average of the vector connecting the shoulder joints and the vector connecting the hip joints) is facing along the positive X axis.
We then project the skeleton's joint locations onto a 2D plane (shortly, we will describe how to select the 2D plane).

Next, for the bones of the upper arms, lower arms, upper legs, lower legs, neck, and spine, we compute the global orientation of each bone within the 2D projection plane.
We then rotate the corresponding bones of the character rig so as to match these global orientations.
Using the new joint positions as ARAP handles, we repose the mesh.
When the character rig is reposed this way, the lengths of the character's bones are never foreshortened.
This is an intentional design decision; foreshortening is quite rare in children's drawings \cite{willats2006making}, and we therefore opted for a method of animation that does not introduce it.

To apply root motion, we compute the per-frame root offset of the human actor and
 scale it by the constant ratio of the actor's average leg length to the character’s average leg length. 
The resulting offset is applied to the character rig, moving it horizontally across the screen. 

When projecting the actor's 3D joint locations onto a 2D plane, there are multiple planes from which to choose.
Which plane to select depends upon the motion:
jumping jacks will be most recognizable when projected onto a frontal plane, while the hip hop dance \textit{running man} will be most recognizable when projected onto a sagittal plane. 
In order for the motion to remain recognizable, the choice of projection plane should preserve as much joint position variance as possible. 

We automatically compute the plane as a function of the motion.
After preprocessing the motion data (as described above), we plot the joint positions over the entire motion clip as a point cloud and perform principal component analysis upon it.
The first two principal components define a 2D plane upon which joint position maximally varies.
The third principal component defines a vector normal to this plane. 
We select as the 2D projection plane either the skeleton's frontal plane or sagittal plane, depending upon which has a normal vector with a higher cosine similarity to the third principal component. 

This projection technique will work well when the source 3D motion primarily occurs on a single plane (such as jumping jacks or a cartwheel).
However, some motions do not cleanly fall onto a single plane, and are therefore more difficult to recognize after being projected to 2D.

To increase the number of motions that remain recognizable, we do not restrict ourselves to using the same 2D plane for the entire skeleton.
Rather, we independently create joint point clouds, perform principal component analysis, and select the projection plane for the upper limbs and the lower limbs (see Figure \ref{fig:rigged_character_creation}, right). 

Mixing perspectives in this manner can result in unrealistic motions, but there can be artistic reasons to justify such deviations from realism\,\cite{Singh:FP:2002}.
Many children's drawings already employ the technique of \textit{twisted perspective}, drawing different parts of a human figure from different points of view\,\cite{dziurawiec1992twisted}.
As a result, mixing perspectives when retargeting matches the motion style to the drawing style, 
increasing the appeal of the final animation as we demonstrate with a user study (see Section \ref{sec:twisted_perspective_perceptual_study}).

\subsection{User Interface}
\label{sec:UI}

\begin{figure*}[ht]
  \centering
  \includegraphics[width=\linewidth]{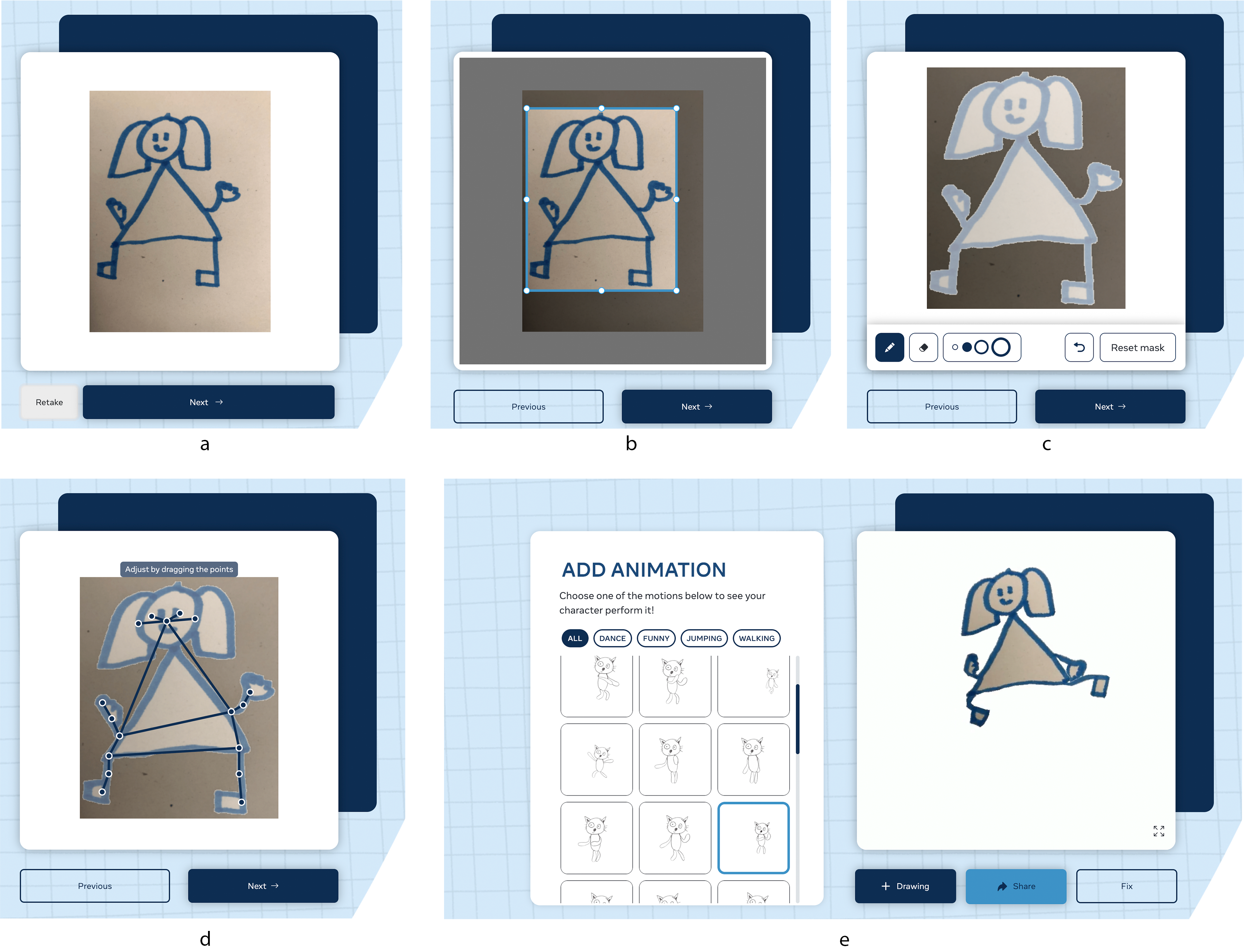}
  \caption{User interface within the \AD Demo. After uploading a drawing (a), users can observe and optionally modify the predicted bounding box (b), modify the segmentation mask (c), and reposition joints (d) prior to selecting a motion to apply to the character rig (e).}
  \Description{fig:UI}
  \label{fig:UI}
  
\end{figure*}

The purpose of the system is to empower users to create appealing animations from their children's drawings.
To increase the chance of a successful outcome, we make certain assumptions about the input image and expose a simple user interface that allows for step-by-step corrections, if needed.

For the detection step, we assume a single human figure to be present within the scene.
If multiple human figures are detected by the model, we return a single bounding box encompassing all detected bounding boxes. 
If no human figures are detected, we return a single bounding box containing the entire image.
Users are prompted to drag the edges of the bounding box to fit to their human figure as needed before continuing to the segmentation step (Figure \ref{fig:UI}.b).

In the segmentation step, users are presented with a visualization of the segmentation mask overlaid on the original image. 
Users can use a pencil and eraser tool to add and subtract pixels from the mask (Figure \ref{fig:UI}.c).
After the user has modified the mask, it is again flood filled and the largest polygon is retained to ensure the mask is a single, solid region (steps e and f in Figure \ref{fig:segmentation-flowchart}).

In the pose detection step, users are shown the predicted joint locations overlaid upon their drawing. 
If a joint is incorrectly positioned, the user can drag it to a more appropriate location (Figure \ref{fig:UI}.d).
Users cannot add or delete joints, but are instructed to drag joints far away from the human figure to avoid using them for animation.

Finally, users are shown a gallery of preselected motions performed by an example character;
clicking on a motion applies it to the user's character rig (Figure \ref{fig:UI}.e).
The gallery of preselected motions is static and does not vary depending upon the uploaded image or annotations.

\hjs{The demo is deployed on Amazon Web Services using a combination of g4dn.2xlarge and c5.4xlarge servers.
If the user makes no annotation modifications, the entire image-to-animation user flow takes less than 10 seconds.}

\section{Evaluation and Analysis}
We evaluate our system in three ways.
First, we briefly describe the public reception of the demo.
Second, we present a set of experiments exploring the effect of training data size on the system's success rate.
Third, we perform a user study to validate the appeal and desirability of twisted perspective motion retargeting.

\subsection{Public Reception}
On December 16, 2021, a version of the proposed system was publicly released as the \AD Demo\,\cite{animateddrawings}.
The launch was accompanied by several high-profile social media posts and a blog post; however, all subsequent online promotion
came from users organically sharing the demo within their networks.

Over the next nine months, over 3.2 million unique users visited the site and spent, on average, over five minutes using the demo.
They uploaded 6.7 million images and, on average, generated four animations per image.
Based upon a subset of highly visible social media posts, the demo is especially popular among parents, elementary school teachers, technology enthusiasts, and artists.

\subsection{Effect of Training Sample Size}
\label{sec:effect_of_training_sample_size}

Our system incorporates repurposed computer vision models trained on photographs of real-world objects.
Because the domain of children's drawings is significantly different in appearance, these models must be fine-tuned prior to use.
However, given the abstract and varied nature of the drawings, it is not obvious how many drawings must be collected and annotated for training.
Therefore, we present a set of experiments exploring the relationship between training dataset size and model prediction success.

We report the performance of the models in two ways. 
Because the models employed have pre-established accuracy metrics, we first report the achieved mean average precision (mAP) [.5:.95] for each model.
However, our goal--animation--is a somewhat distinct downstream use of these predictions, and the mAP may not fully reflect the rate of success.
For example, a predicted bounding box that overlaps ground truth by 90\% would contribute to a very high mAP; 
however, if the prediction excluded a figure's foot or cut off half of its head, the resulting animation would be considered a failure.
Therefore, we also report the percent of predictions that result in successful animations, as determined by visual inspection.

\hjs{We compare the performance of several different fine-tuned versions of our models.
First, we fine-tune using 177,666 images from the Amateur Drawings Dataset; we excluded 500 images to use for validation (as described below).
However, some of the user-accepted annotations are noisy and inaccurate; therefore, we also fine-tune models using `clean' training datasets of multiple sizes.}
To obtain these, we randomly selected and manually reviewed images and annotations from within the Amateur Drawings Dataset.
Images that had clearly incorrect annotations were rejected.
Common reasons for rejection included: segmentation masks that did not contain the entire figure or included background elements, 
limbs that were fused together, joints that did not lie on the figure. 
In this way, we identified 2,500 images with suitable user-accepted annotations to serve as our training and validation data.

We randomly selected 500 images to serve as the validation set across all training runs, while the remaining 2,000 served as the training sample pool.
We created eight different training sets, varying in size from 10 through 2,000. 
For each training set, we randomly selected data samples from the training sample pool of 2,000 until we obtained the appropriately sized set.

We used the model architecture and training parameters specified in Sections \ref{sec:character_detection} (for both detection and segmentation predictions) and \ref{sec:joint_detection}.
Because our goal is to show the effect of training sample size, rather than optimize absolute accuracy, we restrict ourselves to a single model architecture and keep all hyperparameters constant.

To evaluate the percent of predictions suitable for animation, we used the same training sets as described above, but also included the additional set of all 2,500 images. 
For evaluation, we randomly selected an additional 571 images that were uploaded to the \AD Demo. 
While we reviewed these images to ensure that their contents were suitable for animation, we did not review, nor do we make use of, their user-approved annotations. Instead, model predictions were visually inspected to determine whether they would result in a successful animation.

This evaluation was meant to give an assessment of the models' in-the-wild success rates, and not have it be biased towards simpler drawings that our system could already predict perfectly, or those that took little effort to manually correct.
A detection was classified as failure if it did not detect the human figure, detected it multiple times, falsely detected non-human figures in the scene, had a bounding box that cut off a portion of the figure necessary for animation (such as an arm or foot), or had a bounding box extending to include other markings that were not a part of the figure.
A segmentation was classified as failure if it included background elements that were not part of the figure, did not tightly confirm to the bounds of the figure, contained holes in the interior of the figure, was more than one distinct polygon, or connected figure limbs at locations without a joint.
A pose estimation was classified as failure if the nose, shoulders, hips, elbows, knees, wrists, or ankles were not located on or in close proximity to the correct body part.

\subsubsection{Results}
Validation set mAP as a function of fine-tuning training set size is shown in Table \ref{table:mAP}. 
\hjs{Using a Linux server with two NVIDIA Quadro GP100 graphics cards, models trained with 177,666 samples converged in 20 hours, whereas the smaller training sets all converged in under 5 hours.}
For comparison, we also show the mAP obtained when using pretrained model weights (essentially, a fine-tuning training set size of zero) and considering the drawn human figures to be instances of the \textit{person} object class.

\begin{table}[ht]
\resizebox{\columnwidth}{!}{
\begin{tabular}{r|ccc}
\hline
Fine-Tuning &Bounding Box & Segmentation & Pose Estimation\\
Training Set Size & mAP & mAP & mAP \\
\hline
(no fine-tuning) 0 & 0.06 & 0.04 & 0.09 \\
10 & 0.27 & 0.30 & 0.34 \\
100 & 0.51 & 0.51 & 0.76 \\
250 & 0.58 & 0.57 & 0.80 \\
500 & 0.69 & 0.63 & 0.82 \\
1000 & 0.77 & 0.68 & 0.84 \\
1500 & 0.80 & 0.70 & 0.85 \\
2000 & 0.81 & 0.71 & 0.85 \\
\hjs{(noisy) 177,666} & \hjs{0.82} & \hjs{0.49} & \hjs{0.90} \\
\hline
\end{tabular}}
\caption{
Per stage final mAP obtained on validation set as a function of fine-tuning training set size.
} 
\label{table:mAP}
\end{table}

The percentage of successful, animation-ready model predictions on the random 571 test images are given in Table \ref{table:animation_success_rate}.
We report the percentage of predictions that were successful in each stage, as well as the percentage of images for which predictions in all three stages were successful.
Because our system uses the image processing-based approach described in Section \ref{sec:character_segmentation}, we also evaluate this technique's performance using the same segmentation success-failure criteria; 42.4\% of segmentation masks obtained this way were successful.
In parentheses in the rightmost column of Table \ref{table:animation_success_rate}, we report the percentage of images for which predictions in all three stages were successful when the image processing-based segmentation algorithm is used instead of a fine-tuned model prediction.

\begin{table*}
\begin{tabular}{r|cccccc}
\hline
Fine-Tuning       & Bounding Box  & Segmentation  & Segmentation  & Pose Estimation  & All Stages & All Stages\\
Training Set Size & Success Rate  & Success Rate  & Success Rate  & Success Rate     & Success Rate & Success Rate\\
                  &               & \textcolor{gray}{(Mask R-CNN)}  & \textcolor{gray}{(Image Process)}  &            & \textcolor{gray}{(Mask R-CNN Seg.)} & \textcolor{gray}{(Image Process Seg.)}\\
\hline
(no fine-tuning) 0& 0.4  & 0.0  & |    & 0.6  & 0.0  & 0.0\\
                10 & 27.1 & 0.4  & |    & 2.1  & 0.0  & 0.9\\
               100 & 60.9 & 6.4  & |    & 54.1 & 4.9  & 19.4 \\
               250 & 62.2 & 8.2  & |    & 69.5 & 6.4  & 24.0 \\
               500 & 74.4 & 14.5 & 42.4 & 77.4 & 12.8 & 30.5 \\
              1000 & 83.0 & 19.4 & |    & 83.0 & 17.7 & 34.7 \\
              1500 & 89.8 & 20.3 & |    & 87.4 & 19.1 & 37.7 \\
              2000 & 91.8 & 22.7 & |    & 89.5 & 21.2 & 38.9 \\
              2500 & 92.5 & 24.7 & |    & 90.2 & 23.3 & 39.4 \\
\hjs{(noisy) 177,666} & \hjs{92.5} & \hjs{16.1} & |    & \hjs{94.6} & \hjs{16.1} & \hjs{40.6} \\
\hline
\end{tabular}
\caption{
Percentage of model predictions that can successfully be used for animation, as a function of model fine-tuning training set size.
We report the successes per stage for the bounding box, segmentation mask (both Mask R-CNN and image processing-based), and pose estimation predictions.
In the two right-most columns, we report the percentage of images for which the bounding box, segmentation mask, and pose estimation model predictions were all successful. 
} 
\label{table:animation_success_rate}
\end{table*}

\begin{figure*}[ht]
  \centering
  \includegraphics[width=\linewidth]{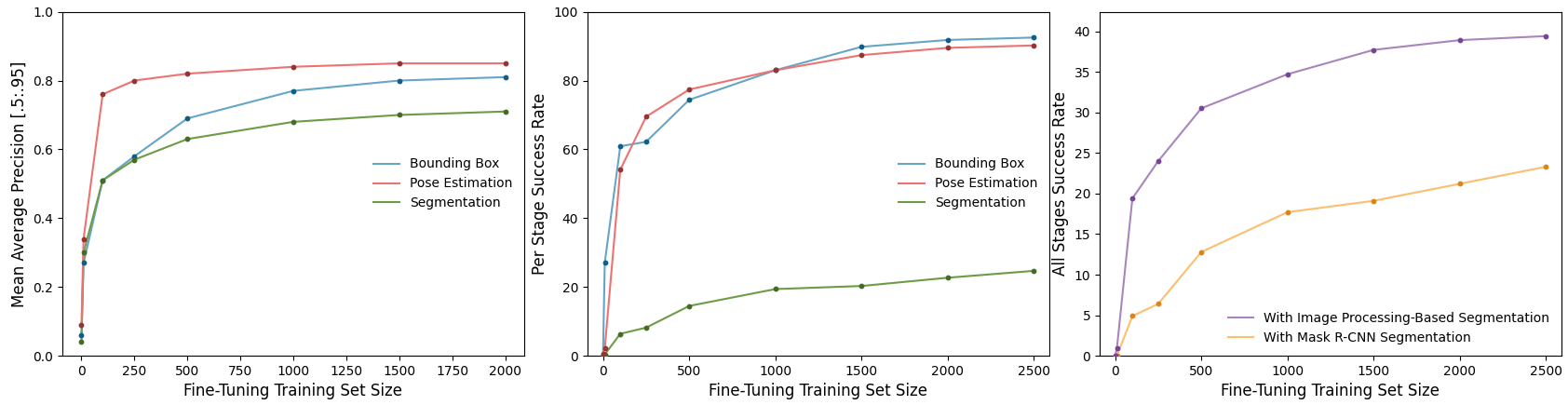}
  \caption{
  \textit{Left:} Achieved mean average precision of bounding box, segmentation, and pose estimation predictions as a function of fine-tuning dataset size.
  \textit{Middle:} Percentage of bounding box, segmentation mask, and pose estimation predictions that could be used for animation without manual correction, respectively.
  \textit{Right:} Percentage of images for which bounding box, segmentation mask, and pose estimation predictions could all be used for animation without manual correction. We show the percentages when using both the Mask R-CNN segmentation predictions and the image processing-based segmentation technique described in Section \ref{sec:character_segmentation}.
  }
  \label{fig:plots}
\end{figure*}

\subsubsection{Discussion}
As Table \ref{table:mAP} shows, directly using model weights trained on real-world images results in very low mAP scores for bounding box, segmentation masks, and pose estimation predictions upon children's drawings.
However, fine-tuning results in a large gain in accuracy across all steps.
Continuing to increase the number of fine-tuning training samples results in continuing, yet slowing, improvements in mAP; 
increasing training set size from 1,500 to 2,000 increases performance by a single percentage point for bounding box and segmentation predictions and does not measurably improve pose estimation.

\hjs{
Interestingly, using the dataset of 177,666 images with noisy annotations results in a minor improvement in bounding box predictions, a significant improvement in pose estimation predictions, and a significant deterioration in segmentation predictions.
Fixing segmentation masks within the \AD Demo is more tedious than fixing bounding box or joint locations.
Therefore, it is likely that more users skipped the segmentation clean-up step, resulting in more noisy segmentation masks within the Amateur Drawings dataset, which in turn lowered the performance of the fine-tuned models.
This insight suggests that, depending upon the complexity of the prediction clean-up tasks offloaded onto the user during data collection, it may or may not be worthwhile to perform additional processing and refinement upon the collected annotations.
}

\hjs{Table \ref{table:animation_success_rate} shows the percentage of model predictions that could successfully be used for animation.
Similarly, without fine-tuning only a very small percentage of bounding box and pose estimation predictions are usable; none of the segmentation predictions are usable.
When fine-tuning with 2,500 `clean' samples, the percentages of usable bounding boxes, segmentation masks, and pose estimations increase to 92.5\%, 24.7\%, and 90.2\%, respectively.
When using the noisy training set of 177,666 images, the pose estimation success rate increased to 94.6\%, while the bounding box success rate was unchanged and the segmentation success rate dropped substantially.
} 
In the supplemental materials, we present many examples of successful and unsuccessful detection and pose predictions from the models trained with 2,500 samples.

Segmentation mask predictions, by contrast, require many more training samples to obtain comparable rates of success;
by a large margin, this step is the most difficult and failure-prone.
With even 2,500 training samples, fewer than one quarter of predictions from Mask R-CNN are suitable for animation without some sort of manual clean-up.
In part, this can be attributed to the presence of many `hollow' or `outline' figures within the dataset, for which the texture of the figure and the texture of background are identical. 
A hollow character's predicted segmentation mask often contains holes within the sparse, non-detailed parts of the figure, and includes connections between non-attached body parts that are drawn close together.
Model predictions also often fail on stick legs and stick arms, which are often missed, especially when other parts of the figure are 2D regions with area.
We present examples of all of these types of failures in the supplemental material.

In comparison, our image processing-based segmentation approach results in a 42.4\% success rate. 
While this approach does a better job of following the outline of the figure, it frequently fails on images with hard shadows introduced during the photographing of the drawing, drawings on lined paper, and figures that are not watertight or have limbs that do not connect.
With the image processing-based segmentation approach, 39.4\% of figures could be fully automatically animated without any manual intervention.
Clearly, further work on robustly segmenting hand drawn figures, or automatically refining the segmentation masks, would be useful in improving the overall success rate.

\subsection{Twisted Perspective Animation Retargeting}
\label{sec:twisted_perspective_perceptual_study}
We evaluate our use of twisted perspective retargeting through a perceptual user study on Amazon Mechanical Turk with 66 subjects.
Subjects were shown a set of 20 videos: four figures that were successfully detected, segmented, and rigged by our system, each performing five different motions (see top of Table~\ref{perceptual_study_results_table}).
Within each video were two side-by-side animations: one animation had been created with twisted perspective, by projecting the lower body and upper body onto different planes, while the other animation used only a single plane of projection.
The side upon which the twisted perspective condition appeared was randomized. 
Both animations played simultaneously, and viewers were asked to select, in a forced-choice manner, the animation whose character motion was `more appealing.' 
To ensure subjects paid attention, four `filter' questions were embedded in the stimuli, in which workers were explicitly directed to select either the left or the right animation.

We present the results in Table~\ref{perceptual_study_results_table}.
For each character and each motion type, we report the percentage of viewers who preferred the animation with twisted perspective motion retargeting over a single perspective.
In parentheses we report significance as the result of a binomial test comparing the distribution of responses to random chance.

In 16 of the 20 videos, a significant preference for twisted perspective was observed. 
In the remaining four videos, there was no significant preference for either type.
Taken together, this result shows that, for these character and motion combinations, twisted perspective retargeting often results in more preferable animation. 
\hjs{Interestingly, three of the four videos in which users had no significant preference depicted figures performing the `Wave Hello' motion.
As can been in the supplemental video, there is significantly less bending of the legs in the `Wave Hello' motion relative to the other motions tested; as a result, twisted perspective retargeting and single perspective retargeting result in more similar character poses.
This observation suggests that twisted perspective retargeting may not be necessary in all situations; rather it is more useful when both the arms and the legs have substantial motion in different planes.
}

\begin{table}[ht]
  \centering
  \includegraphics[width=\linewidth]{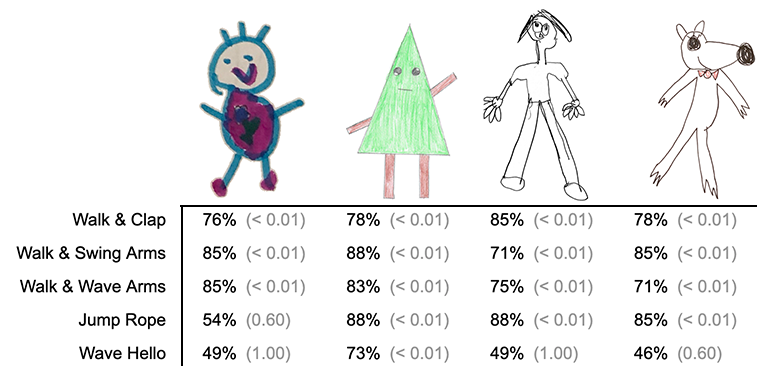}
  \caption{The results of our perceptual study on the use of twisted perspective when retargeting motion. For each character and motion type, we show the percentage of viewers who preferred twisted perspective retargeting and the p-value indicating difference from random chance.}
  \label{perceptual_study_results_table}
  
\end{table}

\section{Amateur Drawing Dataset}
As part of the Animated Drawing Demo, users were asked to consent to a data usage agreement, allowing their uploaded image and annotations to be used for research purposes, including release as part of a public dataset.
Consenting was optional, and refusal to do so did not restrict the experience in any way.
\hjs{Images collected prior to April 20th, 2022 were considered for inclusion into the Amateur Drawings Dataset.
By that date, site users had uploaded over 3.5 million images and consented to the data usage agreement for 1.7 million images.}

\subsection{Refinement}
Many of the images uploaded to the site were photographs of actual people, pets, anime characters, brand logos, and other out-of-domain content.
Therefore, submitted images needed to be filtered to ensure they contained amateur drawings.
This refinement was performed in two steps. 
First, a self-supervised clustering approach was used to identify and filter out-of-domain images.
Second, the remaining images were manually reviewed to ensure their suitability.
\subsubsection{Cluster-based Filtering}
\label{sec:cluster_filtering}
\begin{figure}[ht]
  \centering
  \includegraphics[width=\linewidth]{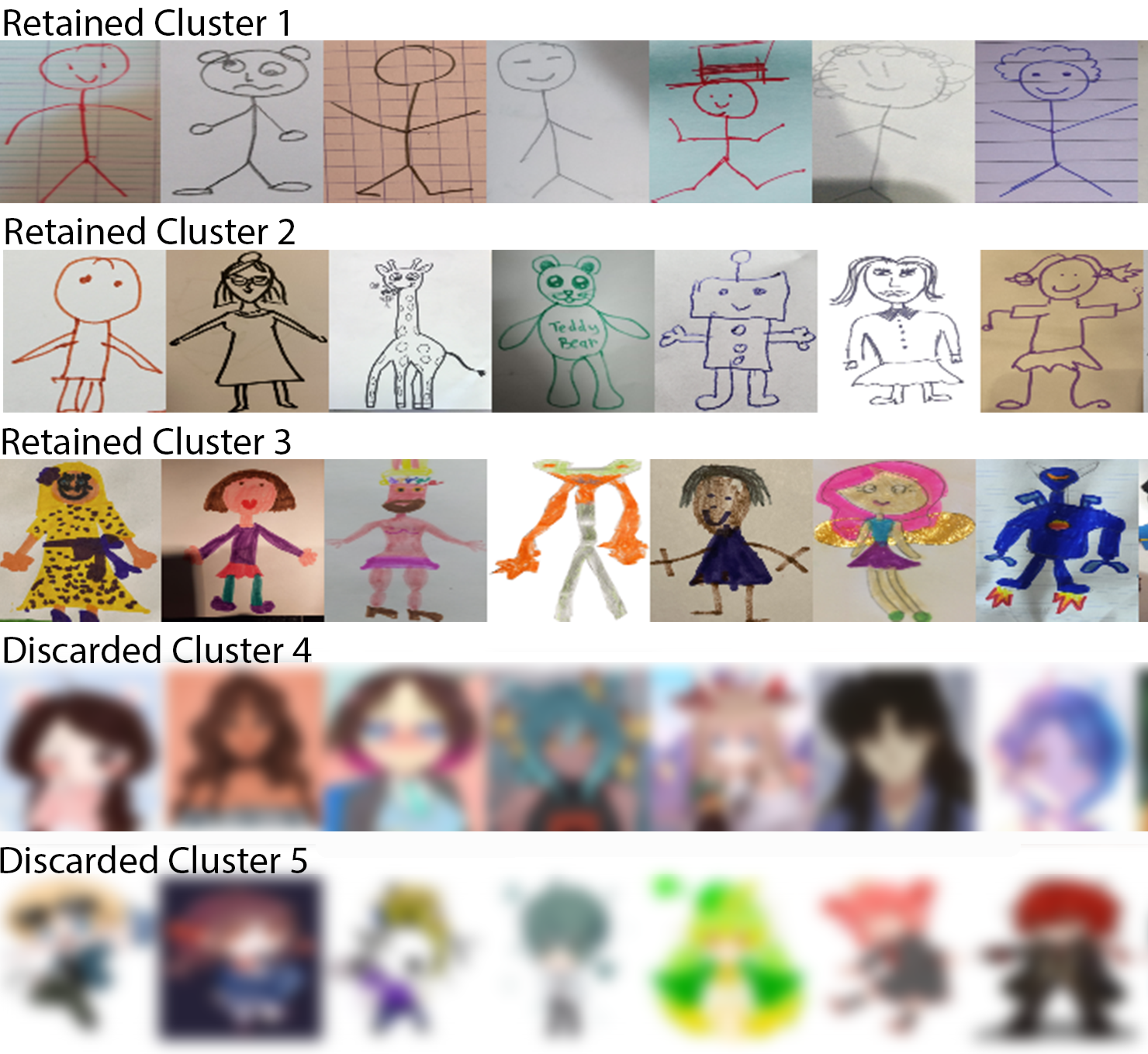}
  \caption{Example images from the clusters obtained via cluster-based filtering. Certain clusters contained similarly depicted characters, such as stick figures, hollow characters, and solid marker characters (retained clusters 1, 2, and 3, respectively). Other clusters contained out-of-domain images, such as anime faces or anime full-body characters (discarded clusters 4, 5 respectively).}
  \Description{fig:clustering_examples}
  \label{fig:clustering_examples}
  
\end{figure}

A self-supervised approach \cite{chen2020improved} was used to train a ResNet-50 feature extractor specific to the consent images.
The feature extractor took the image contents of the figure bounding box and projected it onto a 2048-dimensional embedding space.
Within this space, k-means was used to cluster the embeddings into 100 separate clusters.
From visual inspection, 68 clusters contained out-of-domain subjects, while the remaining 32 clusters primarily contained images of amateur, hand-drawn characters, suitable for inclusion (see Figure \ref{fig:clustering_examples}). 

\begin{figure}[ht]
  \centering
  \includegraphics[width=\linewidth]{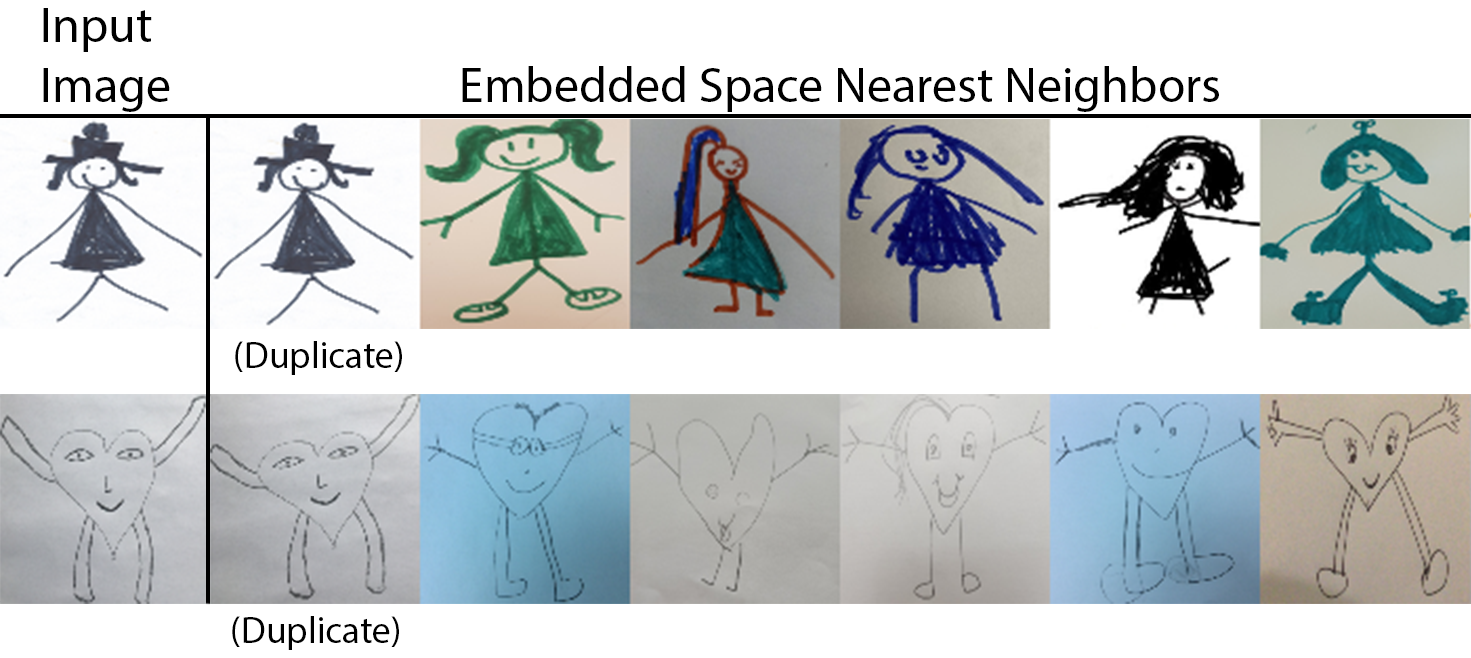}
  \caption{Two input images and their six nearest neighbors within the learned embedding space. As these examples show, duplicates or near-duplicates are quite close within the space, which is useful for filtering them. Similar but distinct figures are also close together within the embedding space; the top figure is close to others with long hair and dresses, while the bottom figure is close to other heart-shaped tadpole figures.}
  \label{fig:embedding_distance}
\end{figure}

Within those 32 clusters were many near-duplicates, images of the same drawing taken from slightly different angles or under slightly different lighting conditions.
Such near-duplicates are close together in the learned embedding space  (see Figure \ref{fig:embedding_distance}).
We detected near-duplicates by computing the Euclidean distance between each pair of images in the embedding space, and removing one of the images if this distance was less than 0.5, a value empirically selected by the authors.
After filtering out-of-domain clusters and removing duplicates, 471,405 images remained.

\subsubsection{Manual Review}
An agency was contracted to review 283,146 of the remaining images.
Reviewers were instructed to ensure images were free-hand, physical drawings containing at least one full-bodied human figure, did not contain characters that are protected intellectual property (such as Mickey Mouse\textsuperscript{TM}), and contained no personally identifiable information or vulgar content. 
Because the reviewers were primarily English speakers, images that contained non-English words were excluded on the basis they might contain inappropriate content.
After manual review, 178,166 images remained.

\hjs{Of the images that were excluded, 
30\% were not freehand drawings, 
24\% did not contain full-bodied human figures, 
20\% contained personally identifiable information, 
15\% contained protected intellectual property, 
4\% were uploaded and annotated with an incorrect orientation,
4\% had out of domain content, 
and 3\% contained vulgar content.}

\subsection{Release}
We are pleased to provide the retained images, along with their annotations, for use by the research community. 
While the \AD Demo was specifically designed for use with children's drawings, the artists' ages were not recorded. 
We therefore refer to the dataset as the Amateur Drawings Dataset.

While the dataset includes the user-accepted character bounding boxes, segmentation masks, and joint positions, we have not attempted to guarantee the accuracy of these annotations.
From a random sampling of 5,000 dataset images and annotations, we observed that
35\% of bounding box detections were modified, 20\% of masks were modified, and  29\% of joint skeletons were modified.
By visual spot check, we confirmed that, in the vast majority of cases, these modifications improved the quality of the annotations.

\section{Conclusion}

In this paper, we present a method to automatically animate the types of drawings created by children and amateur drawers. We also present a first-of-its-kind dataset of 178,166 in-the-wild drawings by children and amateurs, annotated with user-accepted bounding boxes, segmentation masks, and joint locations.

We demonstrate the value of our method in several ways.
First, we explore the accuracy and success from each stage of our system as a function of training dataset size.
Second, we perform a perceptual study to show the appeal of \textit{twisted-perspective} retargeting when animating these characters.
Third, we built and publicly released a usable version of the system which, within its first nine months, has been used to generate over 24 million animations from 6.7 million images uploaded by over 3.2 million users.

\hjs{
Prior to deciding to create a public-facing data collection tool, we unsuccessfully attempted to generate useful synthetic training data using generative adversarial networks\,\cite{pix2pix2017,CycleGAN2017,MultiModalPix2PixNIPS2017_6650}.
We believe our initial collection of less than 1,000 real children's drawings did not contain enough variation to cover the long-tail distribution of the domain. 
In addition, there were many sources of unanticipated nuisance variation that were not in our initial collection, yet present within in-the-wild drawings (e.g., messy backgrounds, lined paper, blurry shots, bad lighting, erased lines). 
It is possible that synthetic data approaches utilizing the entirety of the Amateur Drawing Dataset, which includes these variations, may have more success.

Ultimately, we pivoted to a bootstrapping approach to collect the data necessary to fine-tune our models. 
We manually annotated the images we had and trained initial models, then iteratively released closed beta versions of the demo, collected additional training data, and retrained the models.
By thoughtfully crafting the user experience, keeping prediction and render times short, and providing the user something of value (a downloadable animation of the drawing) in exchange for their efforts, we were able to collect enough real data from our target domain and no longer needed synthetic data.
We would encourage other researchers focused on user-generated content domains, for which there are not yet any suitable datasets, to likewise consider how they might invite their target audience into the dataset creation process, lowering the need to rely upon synthetic data.
}

We believe this work is but a first step towards a robust and comprehensive drawing-to-animation storytelling system, and there are many ways our work could be improved.
One step that can clearly be improved is segmentation.
Extracting a usable and accurate mask can be quite difficult and, because it is used to create the character mesh, even small errors can result in bad animations.
There are many reasons why segmentation is hard to do accurately. 
The photograph of the drawing can be out of focus or distorted due to lighting glare or hard shadows.
Color and texture cues are not guaranteed to be helpful, as in the case of hollow characters.
A line can represent the edge of a body part region, or a line can represent the entire body part, such as with stick figures.
Often, characters are drawn on lined paper, the paper contains eraser marks, or background objects that touch the character are drawn with the same pencil or marker.
\hjs{If the character is drawn with limbs touching in places other than joints (as hands touch hips in the \textit{arms akimbo} pose, for example), there is no predicted segmentation mask that will result in a quality animation unless it is possible to add a segmentation differentiating between the two body parts.

Given the importance and difficulty of the segmentation task, methods that improve the robustness of the masking step would greatly increase the success of our pipeline.
A useful next step could be a principled method for choosing between the image processing and Mask R-CNN segmentation masks on a \textit{per image} basis, as each method can fail for different reasons. 
Ideally, such a method could leverage the bounding box and joint location predictions from the other stages of the pipeline.
}

In addition to improving the robustness of the current pipeline, future work should focus on extracting additional information about the drawing prior to animation. 
A natural next step would be to infer the sub-type of the human figure (e.g., robot, monster, snowman, princess).
Such analysis could be used to modify the pose estimation skeleton (e.g., removing the legs when a snowman has been identified)
or determine the types of animation to apply (e.g., making monsters stomp, princesses dance, or superheroes fly).
It could also be used to infer what different character regions represent.
For example, triangles on a cat's head are ears, while triangles on a devil's head are horns; these insights could affect how the characters are ultimately animated.

\balance  

Many users of the \AD Demo requested, via a feedback form, additional features.
Many wanted support for additional types of motions, or the ability to specify custom motions.
Several requested facial features, such as smiling, blinking, and gaze cues.
Others requested extending the work to handle quadrupeds, multiple characters in a drawing, or to take the context and background of the scene into account when creating the animation.

\hjs{
While our animation method is an appealing way to breathe life into children's drawings, it has two broad limitations.
First, only certain motions can be appealingly retargeted in this manner.
Not all limb motions can be well represented on a 2D plane. 
Spoke-like and arc-like motions, which primarily vary in one or two dimensions, are well handled while carving motions, which vary in all three spatial dimensions, are less recognizable when flattened.
In addition, we always move the character from left-to-right across the page.
If the character is facing right, this should be reversed.
Robustly determining which direction the character is facing is difficult, as the cues may be subtle; for example, the orientation of the nose may be the only cue as to whether the character is facing left or right (see Figure \ref{fig:segmentation_comparison}.h).

Second, our animation method is also limited by the style of the drawing.
We designed the retargeting technique to take advantage of the style of amateur drawings, which lack foreshortening and mix perspective.
If the figure is drawn with foreshortening and proper perspective, the character-motion stylistic mismatch may be undesirable.
In such cases, constructing a proper 3D model of the figure and using a different retargeting technique, such as\,\cite{weng2019photo}, would be preferable.
}

It is our hope that the released dataset will encourage other researchers to focus on methods to analyze and augment amateur drawings.
This domain is a natural form of creativity and expression available to much of the world's population. 
And, given the reception of the \AD Demo, there appears to be widespread appetite for animation and storytelling tools that build upon user-created drawings.

\acknowledgement{
\subsection{Acknowledgement}
We would like to thank FAIR Interfaces, FAIR X, and other members of Meta who helped in the building and release of the demo.
}

\bibliographystyle{ACM-Reference-Format}
\bibliography{Ref}

\end{document}